\documentclass[10pt,twocolumn,letterpaper]{article}

\usepackage[pagenumbers]{cvpr}
\usepackage{graphicx}
\usepackage{amsmath}
\usepackage{amssymb}
\usepackage{booktabs}
\usepackage[table,dvipsnames]{xcolor}
\usepackage{setspace}
\usepackage{comment}
\usepackage{amsmath,amssymb}
\usepackage{color}
\usepackage{wrapfig}
\usepackage{multirow}
\usepackage{floatrow}
\usepackage[normalem]{ulem}
\usepackage{balance}

\usepackage[pagebackref,breaklinks,colorlinks]{hyperref}

\usepackage[capitalize]{cleveref}
\crefname{section}{Sec.}{Secs.}
\Crefname{section}{Section}{Sections}
\Crefname{table}{Table}{Tables}
\crefname{table}{Tab.}{Tabs.}

\def\cvprPaperID{5910} %
\def\confName{CVPR}
\def\confYear{2023}

\begin{document}
{\title{Point Cloud Forecasting as a Proxy for 4D Occupancy Forecasting}

\doublespacing
    \author{Tarasha Khurana\thanks{Equal contribution} \and
    Peiyun Hu$^{*}$ \and
    David Held \and
    Deva Ramanan \and
    Carnegie Mellon University}

\maketitle
}
\begin{abstract}
Predicting how the world can evolve in the future is crucial for motion planning in autonomous systems. Classical methods are limited because they rely on costly human annotations in the form of semantic class labels, bounding boxes, and tracks or HD maps of cities to plan their motion — and thus are difficult to scale to large unlabeled datasets.
One promising self-supervised task is 3D point cloud forecasting~\cite{mersch2022self,wilson2021argoverse,weng2021inverting,weng2022s2net} %
from unannotated LiDAR sequences.
We show that this task requires algorithms to implicitly capture (1) sensor extrinsics (i.e., the egomotion of the autonomous vehicle), (2) sensor intrinsics (i.e., the sampling pattern specific to the particular LiDAR sensor), and (3) the shape and motion of other objects in the scene.
But autonomous systems should make predictions about the world and not their sensors! To this end, we factor out (1) and (2) by recasting the task as one of spacetime (4D) occupancy forecasting. But because it is expensive to obtain ground-truth 4D occupancy, we ``render'' point cloud data from 4D occupancy predictions given sensor extrinsics and intrinsics, allowing one to train and test occupancy algorithms with unannotated LiDAR sequences.
This also allows one to evaluate and compare point cloud forecasting algorithms across diverse datasets, sensors, and vehicles.
\end{abstract}

\section{Introduction}

\begin{figure}
    \centering
    \includegraphics[width=0.5\linewidth]{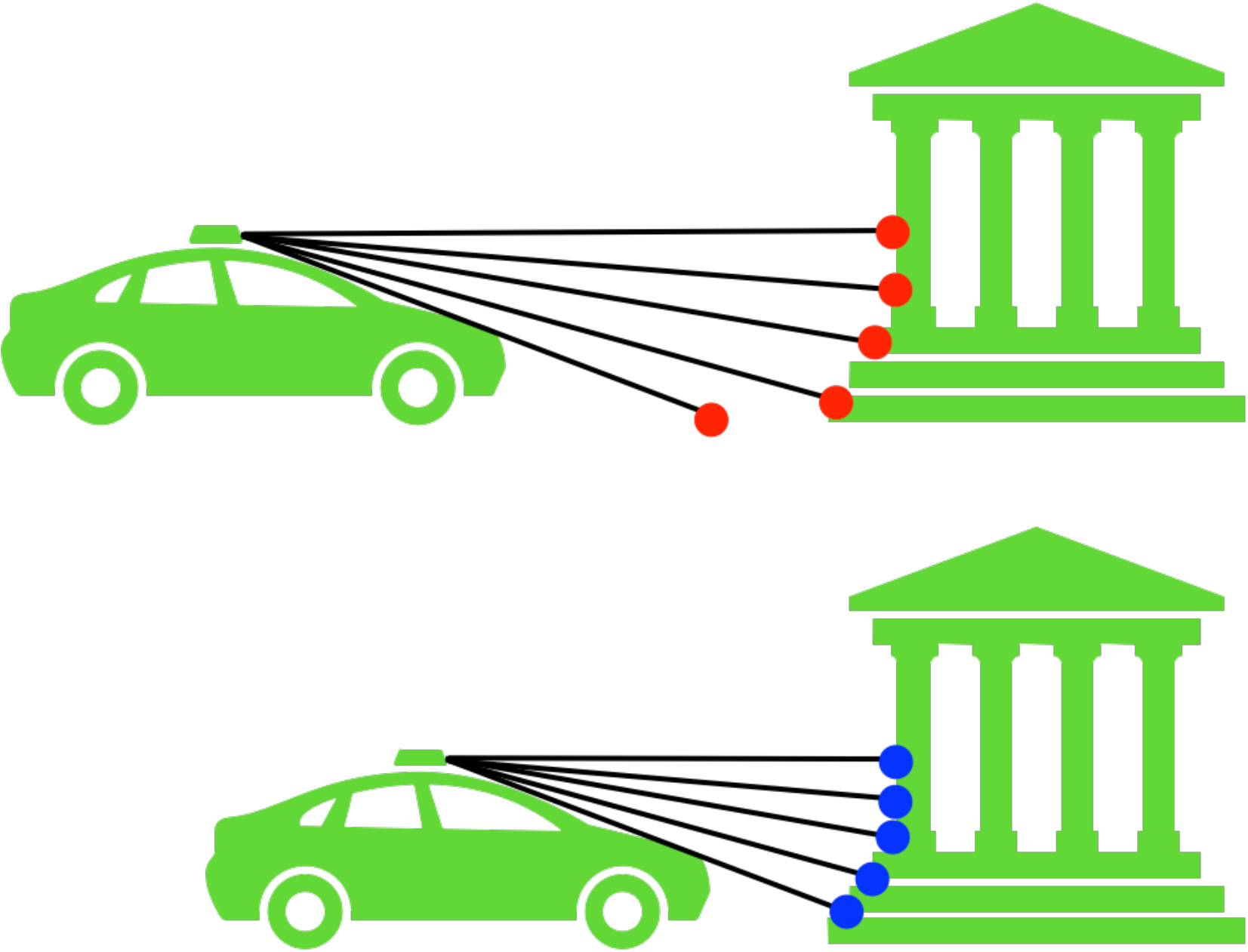}
    \caption{Points depend on the intersection of rays from the depth sensor and the environment. Therefore, accurately predicting points requires accurately predicting sensor extrinsics (sensor egomotion) and intrinsics (ray sampling pattern). But we want to understand dynamics of the environment, not our LiDAR sensor!}
    \label{fig:behind_point_cloud}
\end{figure}

Motion planning in a dynamic environment requires autonomous agents to predict the motion of other objects. Standard solutions consist of perceptual modules such as mapping, object detection, tracking, and trajectory forecasting. Such solutions often rely on human annotations in the form of HD maps of cities, or semantic class labels, bounding boxes, and object tracks, and therefore are difficult to scale to large unlabeled datasets.
One promising \textit{self-supervised} task is 3D point cloud forecasting~\cite{mersch2022self,wilson2021argoverse,weng2021inverting,weng2022s2net}.
Since points appear where lasers from the sensor and scene intersect, the task of forecasting point clouds requires algorithms to implicitly capture
(1) sensor extrinsics (\textit{i.e.}, the ego-motion of the autonomous vehicle),
(2) sensor intrinsics (\textit{i.e.}, the sampling pattern specific to the LiDAR sensor),
and (3) the shape and motion of other objects in the scene.
This task can be non-trivial even in a static scene (Fig.~\ref{fig:behind_point_cloud}).
We argue that autonomous systems should focus on making predictions about the world and not themselves, since an ego-vehicle has access to its future motion plans (extrinsics) and calibrated sensor parameters (intrinsics).

We factor out these (1) sensor extrinsics and (2) intrinsics by recasting the task of point cloud forecasting as one of spacetime (4D) occupancy forecasting. This disentangles and simplifies the formulation of point cloud forecasting, which now focuses solely on forecasting the central quantity of interest, the 4D occupancy.
Because it is expensive to obtain ground-truth 4D occupancy, we ``render'' point cloud data from 4D occupancy predictions given sensor extrinsics and intrinsics.
In some ways, our approach can be seen as the spacetime analog of novel-view synthesis from volumetric models such as NeRFs~\cite{mildenhall2020nerf}; rather than rendering images by querying a volumetric model with rays from a known camera view, we render a LiDAR scan by querying a 4D model with rays from known sensor intrinsics and extrinsics.
This allows one to train and test 4D occupancy forecasting algorithms with un-annotated LiDAR sequences.
This also allows one to evaluate and compare point cloud forecasting algorithms across diverse datasets, sensors, and vehicles.
We find that our approach to 4D occupancy forecasting, which can also render point clouds, performs drastically better than SOTAs in point cloud forecasting, both quantitatively (by up to 3.26m L1 error, Tab.~\ref{tab:reeval}) and qualitatively (Fig.~\ref{fig:qual}). Our method beats prior art with zero-shot cross-sensor generalization (Tab.~\ref{tab:kittiodo}). To our knowledge, these are first results that generalize across train/test sensor rigs, illustrating the power of disentangling sensor motion from scene motion.

\section{Related Work}

\paragraph{Point Cloud Forecasting} As one of the most promising self-supervised tasks that exploit unannotated LiDAR sequences, point cloud forecasting~\cite{mersch2022self, weng2021inverting, weng2022s2net, wilson2021argoverse} provides the algorithm past point clouds as input and asks it to predict future point clouds as output. Traditionally, both the input and the output are defined in the sensor coordinate frame, which moves with time. Although this simplifies preprocessing by eliminating the need for a local alignment, it forces the algorithm to implicitly capture (1) sensor extrinsics (i.e., the egomotion of the autonomous vehicle), (2) sensor intrinsics (i.e., the sampling pattern specific to the particular LiDAR sensor), and (3) the shape and motion of other objects in the scene. We argue that autonomous systems should make predictions about the world and not their sensors. In this paper, we reformulate point cloud forecasting by factoring out sensor extrinsics and intrinsics. Concretely, the new setup asks the algorithm to estimate the depth for rays from future timestamps. We show that one could use it as a proxy for training and testing 4D occupancy forecasting algorithms. Moreover, we demonstrate that one can evaluate existing point cloud forecasting methods under this setup, allowing 4D occupancy forecasting algorithms to be compared with point cloud forecasting algorithms.

\paragraph{Occupancy Forecasting} Occupancy, as a predictive representation complementary to standard object-centric representations in the context of supporting downstream motion planning, has gained popularity over the last few years due to its efficiency in representing complex scenarios and interactions. Most existing works on occupancy forecasting focus on {\em semantic} occupancy grids from a bird's-eye view (BEV)~\cite{sadat2020perceive, casas2021mp3, mahjourian2022occupancy}. They choose to focus on 2D for a good reason since most autonomous driving planners reason in a 2D BEV space. A downside is that it is expensive to obtain ground-truth {\em semantic} BEV occupancy for training and testing algorithms. \cite{khurana2022differentiable} claim that if we reduce our goal from {\em semantic} occupancy to {\em geometric} occupancy, that is knowing if a location is occupied without asking which type of object is occupying it, one could learn to forecast {\em geometric} BEV occupancy from unannotated LiDAR sequences. In this paper, we take the idea from~\cite{khurana2022differentiable} and go beyond BEV -- we propose an approach to learning to forecast 4D {\em geometric} occupancy from unannotated LiDAR sequences. We also propose a scalable evaluation to this task that admits standard point cloud forecasting methods.

\begin{figure*}[h!]
    \centering
    \includegraphics[width=\linewidth]{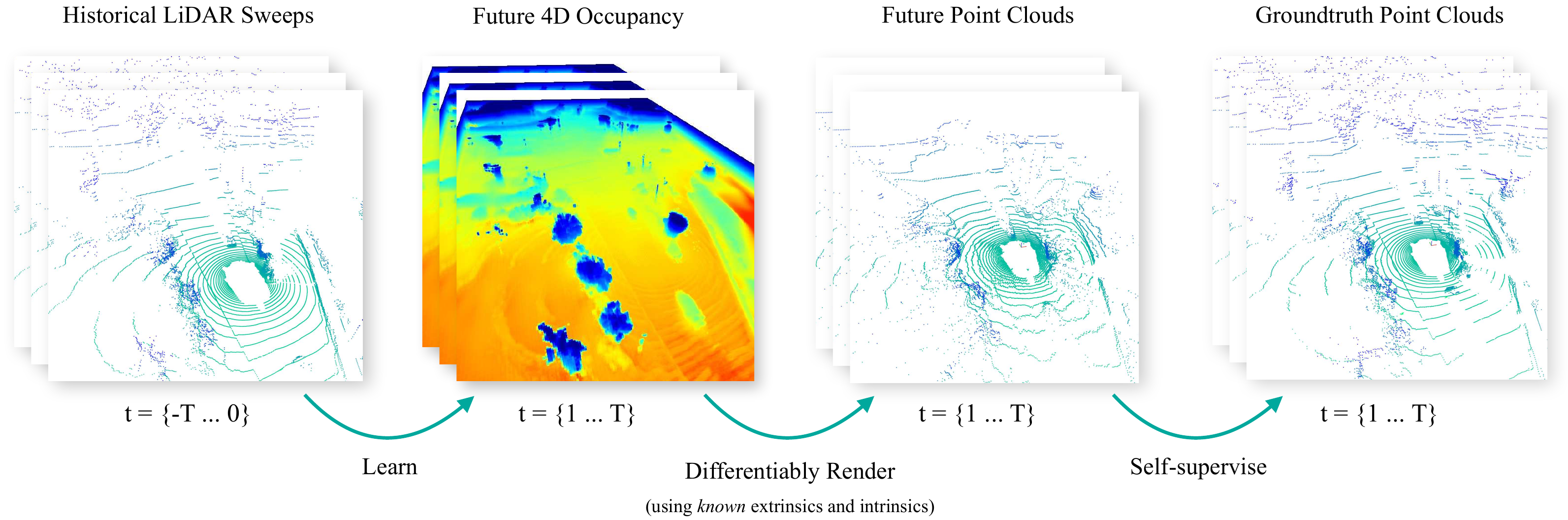}
    \caption{High-level overview of the approach we follow, closely inspired by a prior work \cite{khurana2022differentiable}. Instead of directly predicting future point clouds by observing a set of historical point clouds, we take a geometric perspective on this problem and instead forecast a generic intermediate 3D occupancy-like quantity within a bounded volume. Known sensor extrinsics and intrinsics are an input to our method, which is different from how classical point cloud forecasting is formulated. We argue that this factorization is sensible as an autonomous agent plans its own motion and has access to sensor information. Please refer to our appendix for architectural details.}
    \label{fig:arch}
\end{figure*}

\paragraph{Novel View Synthesis} We have seen tremendous progress in novel view synthesis in the last few years~\cite{lombardi2019neural,mildenhall2020nerf,niemeyer2020differentiable}. At its core, the differentiable nature of volumetric rendering allows one to optimize the underlying 3D structure of the scene by fitting samples of observations with known sensor poses without explicit 3D supervision. Our work can be thought of as novel view synthesis, where we try to synthesize depth images from novel views at future timestamps. Thanks to motion sensors (e.g., IMU), one can assume that relative LiDAR pose among frames in a log can be reliably estimated. Our work also differs from common novel view synthesis literatures in a few important aspects: (a) we use an efficient feed-forward network to predict the spacetime occupancy volume instead of applying test-time optimization; (b) we optimize an explicit volumetric scene representation (i.e., occupancy grid) instead of an implicit neural scene representation; (c) our approach relies on shape and motion prior learned across diverse scenarios in order to predict what happens next instead of reconstructing based on samples only from a specific scenario.

\section{Method}
\label{sec:method}
Autonomous fleets log an abundance of unannotated sequences of LiDAR point clouds $\mathbf X_{-T\dots T}$
, where we also estimate the relative sensor location for each frame $\mathbf o_{-T:T}$.
Suppose we split such a sequence into a historic part $\mathbf X_{-T:0}$ and $\mathbf o_{-T:0}$ and a future part $\mathbf X_{1:T}$ and $\mathbf o_{1:T}$.

Standard point cloud forecasting methods, denoted by function $g$, take the historical sequence of point clouds $\mathbf X_{-T:0}$ as input and try to predict the future sequence of point clouds $\hat{\mathbf X}_{1:T}$.
\begin{equation}
    \hat{\mathbf X}_{1:T} = g(\mathbf X_{-T:0}) \label{eq:pcf1}
\end{equation}

To introduce our approach, we need to first re-parametrize a point from the future LiDAR point cloud, say $\mathbf x \in \mathbf X_t$ where $t=1\dots T$, as a ray that starts from the sensor location $\mathbf o_t$, travels along the direction $\mathbf d$, and reaches the end point $\mathbf x$ after a distance of $\lambda$:
\begin{equation}
    \mathbf x = \mathbf o_t + \lambda \mathbf d, \mathbf x \in \mathbf X_t
\end{equation}

Conceptually, our approach, denoted by function $f$, takes a ray from a future timestamp $t$ parametrized by its origin and direction $(\mathbf o_t, \mathbf d)$, and tries to predict the distance $\hat{\lambda}$ the ray would travel, based on historic sequence of point clouds $\mathbf X_{-T:0}$ and sensor locations $\mathbf o_{-T:0}$.
\begin{equation}
    \hat{\lambda} = f(\mathbf o_t, \mathbf d; \mathbf X_{-T:0}, \mathbf o_{-T:0}) \label{eq:viewsyn}
\end{equation}

Intuitively, Eq.~\eqref{eq:viewsyn} is similar to view synthesis in NERF~\cite{mildenhall2020nerf} except we are computing expected depth rather than expected color. Below, we introduce how we formulate the differentiable volumetric rendering process and use it for learning to forecast 4D occupancy.

\paragraph{Spacetime (4D) occupancy}
We define spacetime occupancy as the occupied state of a 3D location at a particular time instance. We use $\mathbf z$ to denote the true spacetime occupancy, which may not be directly observable due to line-of-sight visibility constraints. Consider a bounded spatialtemporal 4D volume, $\mathcal V$, which is discretized into spacetime voxels $\mathbf v$. We can use
\begin{equation}
    \mathbf z[\mathbf v] \in \{0, 1\}, \mathbf v = (x, y, z, t), \mathbf v\in \mathcal V
\end{equation}
to represent the occupancy of voxel $\mathbf v$ in the spacetime voxel grid $\mathcal V$, which can be {\em occupied} (1) or {\em free} (0).

In practice, we learn an occupancy predictio network $h$ (parametrized by $\mathbf w$) to predict discretized spacetime 4D occupancy given historic sequence of point clouds and sensor locations,
\begin{equation}
    \hat{\mathbf z} = h(\mathbf X_{-T:0}, \mathbf o_{-T:0}; \mathbf w)
\end{equation}
where
\begin{equation}
    \hat{\mathbf z}[\mathbf v] \in \mathbb R_{[0, 1]}
\end{equation}
represents the predicted occupancy of voxel $\mathbf v$ in the spacetime voxel grid $\mathcal V$. Please refer to the appendix for network architecture details.

\paragraph{Depth rendering from occupancy}
\begin{figure}
    \centering
    \includegraphics[width=.7\linewidth]{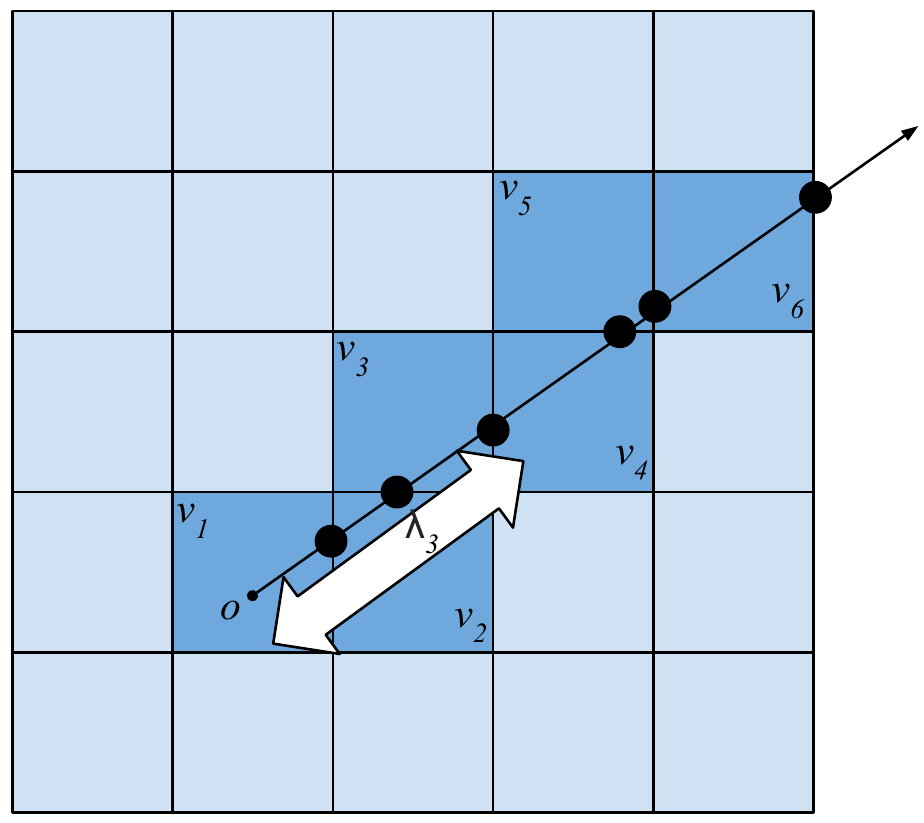}
    \caption{We illustrate the process of rendering depth for a given ray from the predicted occupancy grid. We assume that rays only stop at the voxel boundary, which discretizes the output space into a discrete set of events. We then compute the probability for a ray stopping at each boundary intersection. Finally, we compute the expected stopping distance.}
    \label{fig:volumetric_rendering}
\end{figure}

Given a ray query $\mathbf x = \mathbf o + \lambda \mathbf d$, our goal is to predict $\hat{\lambda}$ as close to $\lambda$ as possible.
We first compute how it intersects with the occupancy grid by voxel traversal~\cite{amanatides1987fast} (Fig.~\ref{fig:volumetric_rendering}).
Suppose the ray intersects with a list of voxels $\{\mathbf v_1\dots \mathbf v_n\}$.
We discretize the ray space by assuming that a ray can only stop at voxel boundaries or infinity.
We interpret occupancy of voxel $\mathbf v_{i}$ as the conditional probability that a ray leaving voxel $\mathbf v_{i-1}$ would stop in voxel $\mathbf v_i$.
We can write
\begin{equation}
    p_i = \prod_{j=1}^{i-1} (1 - \hat{\mathbf z}[\mathbf v_j]) \hat{\mathbf z}[\mathbf v_i]
\end{equation}
where $p_i$ represents the probability that a ray stops in voxel $\mathbf v_i$.
Now we can render the distance by computing the stopping point in expectation.
\begin{equation}
    \hat{\lambda} = f(\mathbf o, \mathbf d) = \sum_{i=1}^n p_i \hat{\lambda}_i \label{eq:render}
\end{equation}
where $\hat{\lambda}_i$ represents the stopping distance at voxel $\mathbf v_i$.

You may have noticed that Eq.~\eqref{eq:render} does not capture the case where the ray stops outside the voxel grid, where the stopping distance is ill-defined (it will stop at infinity). During training, we allow a virtual stopping point outside the grid at the ground-truth location, i.e.,
\begin{equation}
    \hat{\lambda} = f(\mathbf o, \mathbf d) = \sum_{i=1}^n p_i \hat{\lambda}_i + \prod_{i=1}^n (1-p_i) \hat{\lambda}_{n+1} \label{eq:render2}
\end{equation}
where $\hat{\lambda}_{n+1} = \lambda$.

\paragraph{Loss function}
We can train the occupancy prediction network with a simple L1 loss between the rendered depth $\hat{\lambda}$ and the ground-truth depth $\lambda$.
\begin{equation}
    L(\mathbf w) = \sum_{(\mathbf o, \lambda, \mathbf d) \in (X_{1:T}, \mathbf o_{1:T})} \left|\lambda - f(\mathbf o, \mathbf d; \mathbf X_{-T:0}, \mathbf o_{-T:0}, \mathbf w)\right|
\end{equation}

\section{Evaluation}
\label{sec:evaluation}

\begin{figure}
    \centering
    \includegraphics[width=\linewidth]{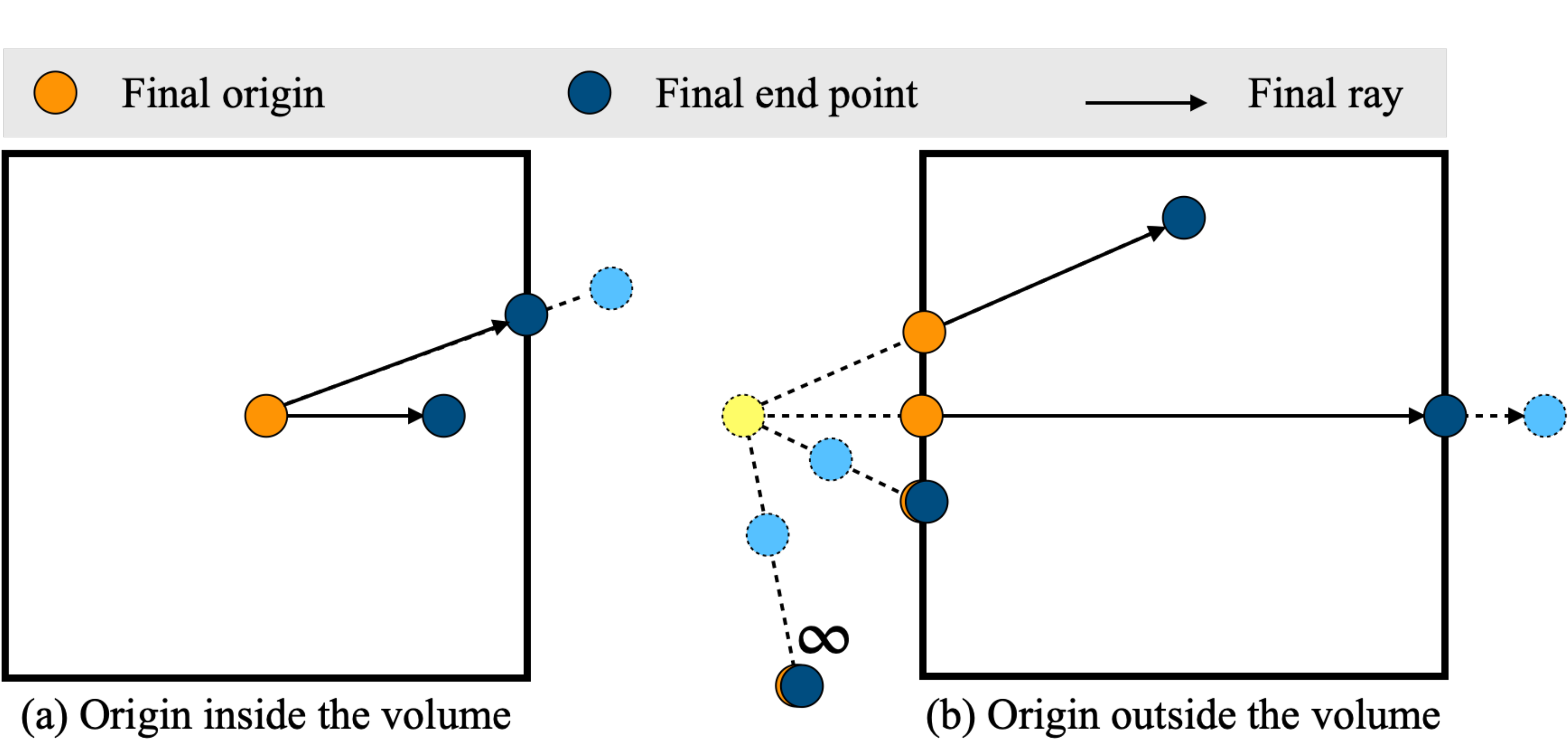}
    \caption{Ray Clamping. First, we move the origin towards the end point until the origin touches the volume or infinity. Then, we move the end point towards the origin until the end point touches the volume or infinity. At all times, we make sure the end point stays ahead of the origin (like two rings on a string). Being inside the volume counts as touching it.}
    \label{fig:clamp}
\end{figure}

The golden standard for evaluating 4D occupancy forecasting would be to compare the predicted occupancy with the ground-truth, but because it is extremely expensive to obtain ground-truth 4D occupancy, we ``render'' future point clouds from forecasted 4D occupancy with known sensor intrinsics and extrinsics, use the quality of rendered future point clouds as a proxy for that of forecasted 4D occupancy.

We introduce a new evaluation, where we factor out sensor intrinsics and extrinsics such that algorithms can be evaluated solely based on how well it captures how the scene unfolds. We provide future rays as queries and ask algorithms to provide a depth estimate for each query.

Given a query ray $\overrightarrow{OQ}$, there is a prediction ray $\overrightarrow{OP}$, where $O$ represents the origin, $Q$ represents the ground-truth end point, and $P$ represents the predicted end point.
\begin{align}
    \overrightarrow{OQ} &= \mathbf o + \lambda \mathbf d\\
    \overrightarrow{OP} &= \mathbf o + \hat \lambda \mathbf d
\end{align}

Given such a pair of rays, we define the error $\varepsilon$:
\begin{align}
   \varepsilon &= |\overrightarrow{OQ} - \overrightarrow{OP}| = |\overrightarrow{PQ}| = |\lambda - \hat \lambda|\label{eq:eval}
\end{align}

\begin{figure*}
    \centering
    \includegraphics[width=\linewidth]{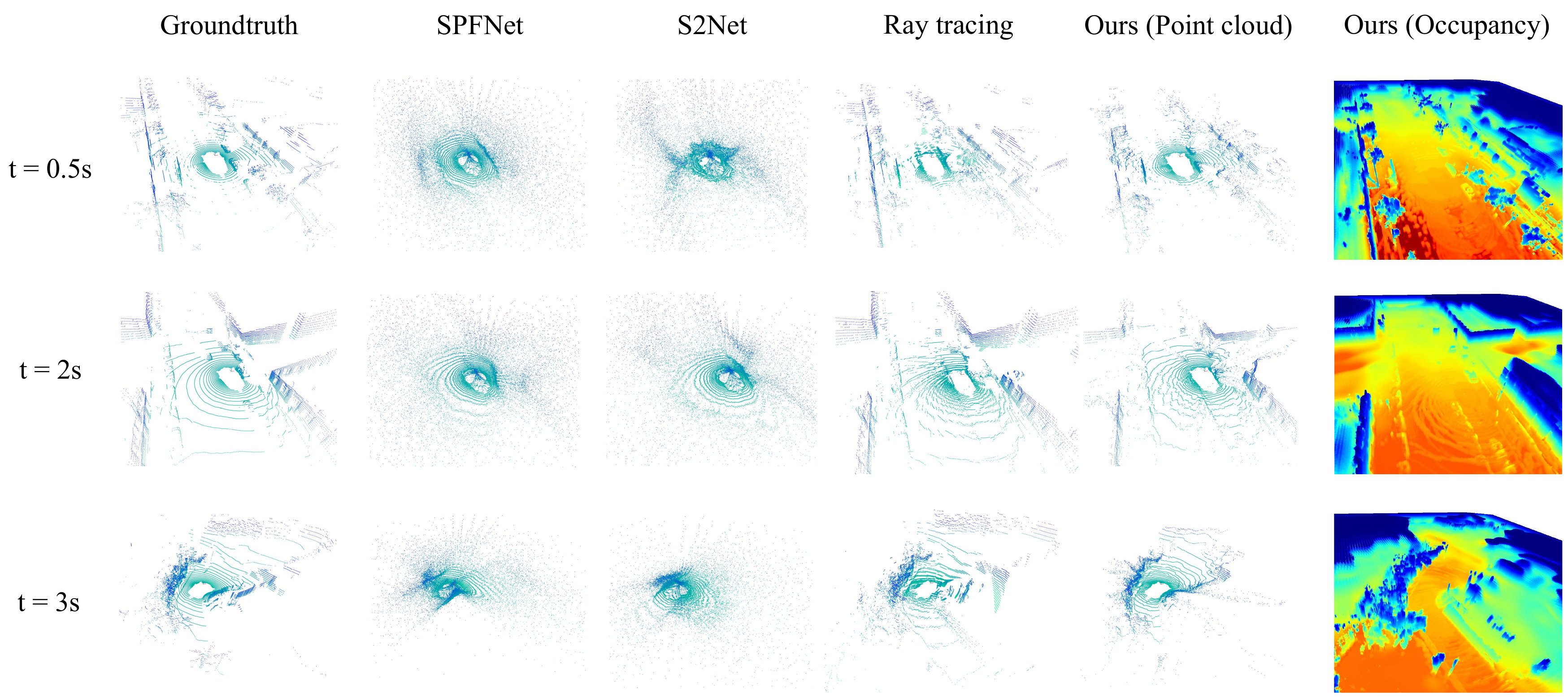}
    \caption{Qualitative results. We compare the point cloud forecasts of S2Net~\cite{weng2022s2net}, SPFNet~\cite{weng2021inverting} and the raytracing baseline on the nuScenes dataset with our approach on three different sequences at different time horizon. Our forecasts look significantly crisper than the SOTA. This demonstrates the benefit of learning to forecast spacetime 4D occupancy with sensor intrinsics and extrinsics factored out. We also visualize the forecasted 4D occupancy at the corresponding future timestamp. As compared to simple \textit{aggregation}-based raytracing, we are able to \textit{spacetime-complete} 4D scenes. We highlight some potential applications in Fig.~\ref{fig:domaintransfer} and Fig.~\ref{fig:densify}. We visualize a render of the predicted occupancy and the color encodes height along the z-axis.
    }
    \label{fig:qual}
\end{figure*}

\paragraph{Near-field error}
Since LiDAR rays only travel through freespace and terminate when reaching occupied surface, there is a physical meaning behind the $\varepsilon$ in Eq.~\eqref{eq:eval}. In practice, occupancy and freespace prediction is only relevant in regions that are reachable by the autonomous vehicle in planning's time hoziron.
To reflect the focus on the reachable regions,
we propose an operation to clamp any given ray $\overrightarrow{XY}$ to the fixed volume $\mathcal{V}$. We call it {\it ray clamping}, denoted as $\phi_{\mathcal{V}}: \overrightarrow{XY} \rightarrow \overrightarrow{X'Y'}$ and illustrated in Fig.~\ref{fig:clamp}.

We define the near-field (bounded by volume $\mathcal{V}$) prediction error $\varepsilon_{\mathcal{V}}$ as
\begin{align}
    \varepsilon_{\mathcal{V}} = |\phi_{\mathcal{V}}(\overrightarrow{OQ}) - \phi_{\mathcal{V}}(\overrightarrow{OP})| = |\overrightarrow{O'Q'}-\overrightarrow{O'P'}|=|\overrightarrow{P'Q'}|
\end{align}

Even though this metric penalizes disagreements of predicted depth along query rays within the bounded volume, it does not capture the severity of a prediction error. In real-world, one meter of an error close to the AV matters more. To this end, we also propose using a relative near-field prediction error $\varepsilon^{rel}_{\mathcal{V}}$ defined as,
\begin{equation}
    \varepsilon^{rel}_{\mathcal{V}} = \frac{|\phi_{\mathcal{V}}(\overrightarrow{OQ}) - \phi_{\mathcal{V}}(\overrightarrow{OP})|}{|\overrightarrow{OQ}|} =\frac{|\overrightarrow{P'Q'}|}{|\overrightarrow{OQ}|}
\end{equation}

The proposed evaluation requires one predicted ray for every ground-truth ray (query). Any algorithms that are capable of rendering depth for a given ray by design meets this requirement, including 4D occupancy forecasting from Sec.~\ref{sec:method}. However, for point cloud forecasting algorithms, the number of predicted points does not necessarily match the number of ground-truth rays, plus there is no one-to-one mapping between predicted and ground-truth points.
To resolve this discrepancy, we propose to fit a surface to the predicted point clouds, on which we can query each ground-truth ray, find its intersection with the fitted surface, and output the (clamped) ray distance. In practice, we interpolate depth among the spherical projections of predicted rays.

We also consider vanilla chamfer distance $d$~\eqref{eq:rawcd} and near-field chamfer distance $d_\mathcal{V}$~\eqref{eq:nfcd}
\begin{equation}
    d = \frac{1}{2N}\sum_{\mathbf{x} \in \mathbf{X}} \min_{\mathbf{\hat{x}} \in \hat{\mathbf{X}}} || \mathbf{x} - \mathbf{\hat{x}} ||_2^2 +
   \frac{1}{2M}\sum_{\hat{\mathbf{x}} \in \hat{\mathbf{X}}} \min_{\mathbf{x} \in \mathbf{X}} || \mathbf{x} - \mathbf{\hat{x}} ||_2^2
    \label{eq:rawcd}
\end{equation}
where $\mathbf X$, $\hat{\mathbf X}$ represents the ground-truth, predicted point cloud; $N$ and $M$ are their respective number of points.
\begin{equation}
d_{\mathcal{V}} = \frac{1}{2N'}\sum_{\mathbf{x} \in \hat{\mathbf{X}_\mathcal{V}}} \min_{\mathbf{\hat{x}} \in \mathbf{X}_\mathcal{V}} || \mathbf{x} - \mathbf{\hat{x}} ||_2^2 + \frac{1}{2M'}\sum_{\mathbf{x} \in \mathbf{X}_\mathcal{V}} \min_{\mathbf{\hat{x}} \in \hat{\mathbf{X}}_\mathcal{V}} || \mathbf{x} - \mathbf{\hat{x}} ||_2^2
    \label{eq:nfcd}
\end{equation}
where $\mathbf X_\mathcal{V}$, $\hat{\mathbf X}_\mathcal{V}$ represents the ground-truth point cloud and predicted point cloud within the bounding volume $\mathcal{V}$; $M'$, $N'$ are their respective number of points.

\section{Experiments}

\paragraph{Datasets} We perform experiments on nuScenes \cite{caesar2019nuscenes}, KITTI-Odometry~\cite{geiger2013vision, behley2019semantickitti} and ArgoVerse2.0 \cite{wilson2021argoverse}. nuScenes \cite{caesar2019nuscenes} is a full-suite autonomous driving dataset with a total of 1,000 real-world driving sequences of 15s each. KITTI \cite{geiger2013vision} is also a multi-sensor dataset with 6 hours of diverse driving data across freeways and urban areas. KITTI-Odometry is a subset of this KITTI dataset where sequences have accurate sensor poses. ArgoVerse2.0~\cite{wilson2021argoverse} contains the largest set of unannotated LiDAR sequences. Please see the appendix for results on ArgoVerse2.0.

\paragraph{Setup} We consider a bounded area around the autonomous vehicle: -70m to 70m in the x-axis, -70m to 70m in the y-axis and -4.5m to 4.5m in the z-axis in the nuScenes coordinate system. This is our 4D volume $\mathcal{V}$, described in Sec. \ref{sec:method}. We follow the state-of-the-art in point cloud forecasting and evaluate forecasting in a 1 second horizon and a 3 second horizon. We adopt the same setup as prior methods~\cite{weng2021inverting, weng2022s2net}. On nuScenes, for 1s forecasting,, we take 2 frames of input and 2 frames of output at 2Hz;
for 3s forecasting, we take 6 frames of input and 6 frames of output
 at 2Hz.
 For all other datasets, we always take 5 frames of input and 5 frames of output for both 1s and 3s forecasting.

 \paragraph{Baselines} First, we construct an aggregation-based raytracing baseline (similar to \cite{mersch2022self}). Specifically, we populate a binary occupancy grid given the aligned LiDAR point clouds from the past and present timesteps and use it for querying ground-truth rays. In addition to this, we compare our 4D occupancy forecasting approach to state-of-the-arts (SOTAs) in point cloud forecasting, including SPFNet~\cite{weng2021inverting} and S2Net~\cite{weng2022s2net} on the nuScenes dataset, and ST3DCNN~\cite{mersch2022self} on the KITTI-Odometry dataset. For SPFNet~\cite{weng2021inverting} and S2Net~\cite{weng2022s2net}, we are able to obtain the raw point cloud predictions from the authors and evaluate the results on the new metrics. For fair comparison, the S2Net results are based on a single sample from their VAE. We retrain ST3DCNN \cite{mersch2022self} models for 1s and 3s forecasting.

In addition, the state-of-the-art approaches (barring ST3DCNN) tend to predict a confidence score for each point, indicating how valid the predicted point is; we %
evaluate the predicted point cloud both with and without confidence filtering, with a recommended confidence threshold at 0.05~\cite{weng2021inverting, weng2022s2net}. Quantitative and qualitative results with confidence filtering can be found in the appendix.

\newpage

\subsection{Re-evaluate state-of-the-arts}
\paragraph{Qualitative results on nuScenes}
\begin{table}[t]
  \centering
  \resizebox{\columnwidth}{!}{%
  \begin{tabular}{@{}cccccc@{}}
  \toprule[1.5pt]
	\multirow{2}{*}{Method} & \multirow{2}{*}{Horizon} & \multirow{2}{*}{L1 (m)} & \multirow{2}{*}{AbsRel (\%)} & \multicolumn{2}{c}{Chamfer Distance ($m^2$)}\\
	& & & & Near-field & Vanilla\\
	\midrule
    \multirow{2}{*}{S2Net \cite{weng2022s2net}} & 1s & 3.49 & 28.38 & 1.70 & 2.75 \\
    & 3s & 4.78 & 30.15 & 2.06 & \textbf{3.47 }\\
    \midrule
    \multirow{2}{*}{SPFNet \cite{weng2021inverting}} & 1s & 4.58 & 34.87 & 2.24 & 4.17 \\
    & 3s & 5.11 & 32.74 & 2.50 & 4.14 \\
        \midrule
    \multirow{2}{*}{Ray tracing} & 1s & 1.50 & 14.73 & \textbf{0.54} & \textbf{0.90} \\
    & 3s & 2.44 & 26.86 & 1.66 & 3.59 \\
    \midrule
        \multirow{2}{*}{Ours} & 1s & \textbf{1.40} & \textbf{10.37} & 1.41 & 2.81 \\
    & 3s & \textbf{1.71} & \textbf{13.48} & \textbf{1.40} & 4.31 \\
    \bottomrule[1.5pt]
  \end{tabular}
  \caption{Results on nuScenes \cite{caesar2019nuscenes}. We see that the conclusions made from the proposed metrics are more in line with the qualitative results in Fig.~\ref{fig:qual}. This reiterates the need for metrics that intuitively evaluate the underlying \textit{geometry} of the scene instead of uncorrelated samples of the scene (e.g., points in space).
  }
  \label{tab:reeval}
}
\end{table}

\begin{table}[t]
  \centering
  \resizebox{\columnwidth}{!}{%
  \begin{tabular}{@{}ccccccc@{}}
  \toprule[1.5pt]
	\multirow{2}{*}{Method} & Train & \multirow{2}{*}{Horizon} & \multirow{2}{*}{L1 (m)} & \multirow{2}{*}{AbsRel (\%)} & \multicolumn{2}{c}{Chamfer Dist. ($m^2$)}\\
	& set & & & & Near-field & Vanilla \\
    \midrule
    \multirow{2}{*}{ST3DCNN \cite{mersch2022self}} & \multirow{2}{*}{KITTI-O} & 1s & 3.13 & 26.94 & 4.11 & 4.51\\
    & & 3s & 3.25 & 28.58 & 4.19 & 4.83\\
    \midrule
    \multirow{2}{*}{Ours} & \multirow{2}{*}{ KITTI-O} & 1s & \textbf{1.12} & \textbf{9.09} & \textbf{0.51} & \textbf{0.61} \\
    & & 3s & \textbf{1.45} & \textbf{12.23} & \textbf{0.96} & \textbf{1.50} \\
    \midrule
    \midrule
    \multirow{2}{*}{Ray tracing} & \multirow{2}{*}{-} & 1s & \textbf{1.50} & 16.15 & \textbf{0.62} & \textbf{0.76} \\
    & & 3s & 2.82 & 29.67 & \textbf{4.01} & 5.92 \\
    \midrule
    \multirow{2}{*}{Ours} & \multirow{2}{*}{AV2} & 1s & 1.71 & \textbf{14.85} & 2.52 & 3.18\\
    & & 3s & \textbf{2.52} & \textbf{23.87} & 4.83 & \textbf{5.79} \\
    \midrule
    \midrule
    \multirow{2}{*}{Ours} & \multirow{2}{*}{$\text{KITTI-O}^{20\%}$} & 1s & 1.25 & 9.69 & 1.95 & 2.27 \\
    & & 3s & 1.70 & 14.09 & 4.09 & 5.09 \\
    \midrule
    \multirow{2}{*}{Ours} & AV2 + & 1s & \textbf{1.19} & \textbf{9.30} & \textbf{0.54} & \textbf{0.64} \\
    & $\text{KITTI-O}^{20\%}$ & 3s & \textbf{1.67} & \textbf{13.40} & \textbf{1.24} & \textbf{1.80}\\
    \bottomrule[1.5pt]
  \end{tabular}
  \caption{
  Performance as a function of the available target dataset (in this case, KITTI-Odometry).
  With access to all of KITTI-O ({\bf top}), our method outperforms the SOTA.
  With no access to KITTI-O (\textit{i.e.} zero-shot sensor generalization in the {\bf middle}), our method trained on AV2 outperforms the ray tracing baseline at 3s, though the baseline fares well at 1s. Note that both approaches still beat the SOTA \cite{mersch2022self} by a large margin.
  Finally, with access to only 20\% of KITTI-O ({\bf bottom}), our method fares quite well, particularly when trained on both AV2 and KITTI-O.
  Cross-dataset generalization and training is made possible by disentangling sensor intrinsics/extrinsics from scene motion.
  }
  \label{tab:kittiodo}
  }
\end{table}

We compare the forecasted point clouds from our 4D occupancy forecasting approach to SOTA on point cloud forecasting in Fig.~\ref{fig:qual}, where we see a drastic difference in how the predicted point clouds look like. Our forecasts look significantly more representative of the scene geometry compared to SOTA.
This demonstrates the benefit of learning to forecast spacetime 4D occupancy with sensor intrinsics and extrinsics factored out. Surprisingly, we find that aggregation-based raytracing is a competitive baseline, qualitatively better than the SOTA. However, in addition to this \textit{aggregation}, our approach is also able to hallucinate or \textit{spacetime-complete} both the future motion of dynamic objects and the occluded parts of the static world.
We also visualize the 3D forecasted occupancy at corresponding timestamps that our approach predicts ``for free''.
Please refer to the caption for more details.

\paragraph{Results on nuScenes with new metrics}

We compare our 4D occupancy forecasting to SOTA on point cloud forecasting in terms of depth error along the future rays, following the evaluation protocol outlined in Sec.~\ref{sec:evaluation}. We find that the 4D occupancy forecasting approach outperforms all baselines by significant margins in both 1s and 3s forecasting, reducing both the L1 and the absolute relative error by more than half, compared to the state-of-the-art methods on point cloud forecasting. The improvements here are consistent with the qualitative results in Fig.~\ref{fig:qual}. As noted before, the raytracing baseline performs better than SOTA.

\paragraph{Results on nuScenes with old metrics}

We also evaluate by both vanilla~\eqref{eq:rawcd} and near-field chamfer distance~\eqref{eq:nfcd} following the protocol in Sec.~\ref{sec:evaluation}.

Our approach shines in terms of near-field chamfer distance. One contributing factor could be that our approach is specifically optimized for capturing occupancy evolution in the near field. In addition, S2Net~\cite{weng2022s2net} outperforms us in terms of vanilla chamfer distance, which is not surprising since we are incapable of deciding where rays end outside the predefined voxel grid.

\paragraph{Results on KITTI-Odometry} Next, we use KITTI-Odometry to test our method in different settings with limited access to the target dataset. This mimics the setting where a next-generation sensor platform may be gradually integrated into fleet operations. Tab.~\ref{tab:kittiodo} shows that with access to the full target dataset (KITTI-Odometry) for training, our method resoundingly outperforms the SOTA ST3DCNN \cite{mersch2022self}.
Next, if no samples from the target dataset are available, one can employ either a non-learnable method such as our raytracing baseline, %
 or one may pretrain on a (large) dataset with a different sensor platform. To this end, we find that our method trained on ArgoVerse2.0 outperforms the SOTA on KITTI-Odometry, while also outperforming raytracing baseline for long-horizon (3s) forecasting. Finally, with access to only 20\% of KITTI-Odometry, our method pretrained on ArgoVerse2.0 and finetuned on KITTI-Odometry outperforms the alternatives. \textit{To our knowledge, these are the first results in sensor transfer/generalization that illustrate the power of disentangling sensor extrinsics/intrinsics from scene motion.} Please see qualitative results in the appendix.

\begin{figure*}[t]
    \centering
    \includegraphics[width=\linewidth]{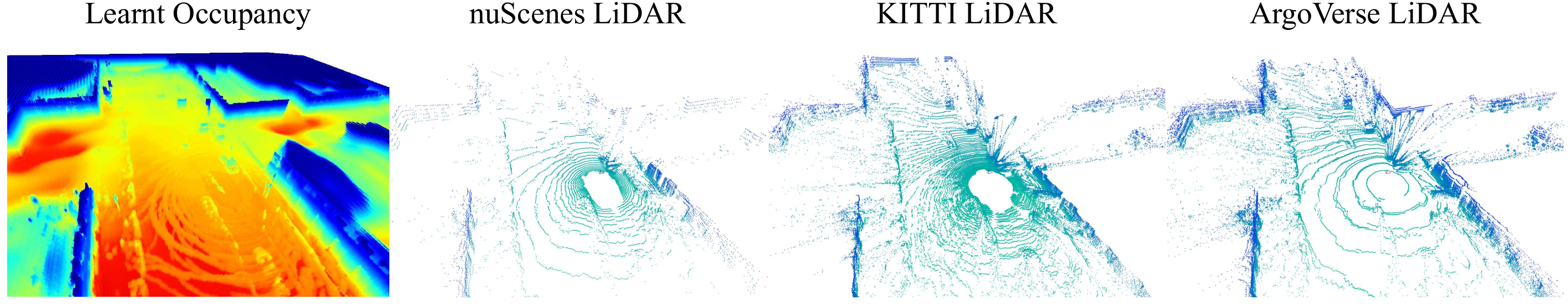}
    \caption{\textbf{Novel intrinsic-view synthesis} We show how to simulate different LiDAR ray patterns on top of the same learned occupancy grid. In this case, the future occupancy is predicted with historic LiDAR data scanned by nuScenes LiDAR (Velodyne HDL32E). First, we show the rendered point cloud under the native setting. Then, we show the rendered point cloud for KITTI LiDAR (Velodyne HDL64E, 2x as many beams). Finally, we have the rendered point cloud for Argoverse 2.0 LiDAR (2 VLP-32C stacked on top of each other). The fact that we can forecast occupancy on top of data captured by one type of sensor and use it to simulate future data for different sensors shows how generic the forecasted occupancy is as a representation. We support this generalization quantitatively in Tab.~\ref{tab:kittiodo}.}
    \label{fig:domaintransfer}
\end{figure*}

\subsection{Architecture ablations}

Here, we explore two other variants of our architecture: a \textit{static} variant that predicts a single voxel grid for all future timesteps, and a \textit{residual} variant that predicts a single static voxel grid with residual voxel grids for each output timestep. We evaluate these variants on nuScenes.

\begin{table}[t]
  \centering
  \resizebox{\columnwidth}{!}{%
  \begin{tabular}{cccccc}
  \toprule[1.5pt]
	\multirow{2}{*}{Arch.} & \multirow{2}{*}{Horizon} & \multirow{2}{*}{L1 (m)} & \multirow{2}{*}{AbsRel (\%)} & \multicolumn{2}{c}{Chamfer Distance ($m^2$)}\\
	& & & & Near-field & Vanilla\\
    \midrule
    \multirow{2}{*}{S} & 1s & \textbf{1.28} & \textbf{9.27} & 1.03 & 3.41\\
    & 3s & 1.73 & 13.54 & \textbf{1.40} & 3.73 \\
	\midrule
    \multirow{2}{*}{D} & 1s & 1.40 & 10.37 & 1.41 & \textbf{2.81} \\
    & 3s & \textbf{1.71} & \textbf{13.48} & \textbf{1.40} & 4.31 \\
    \midrule
    \multirow{2}{*}{S+R} & 1s & 1.34 & 9.73 & \textbf{1.00} & 3.20 \\
    & 3s & 1.82 & 13.84 & 1.52 & \textbf{3.54} \\
    \bottomrule[1.5pt]
  \end{tabular}
  \caption{We evaluate two variants of the proposed dynamic (D) architecture using the geometry forecasting metrics - static (S) and residual (S+R). We find that the static variant is a powerful baseline that beats our dynamic approach for 1s forecasting and by extension, the state-of-the-art. }
  \label{tab:variants}
  }
\end{table}

\begin{figure*}[t]
    \centering
    \includegraphics[width=\linewidth]{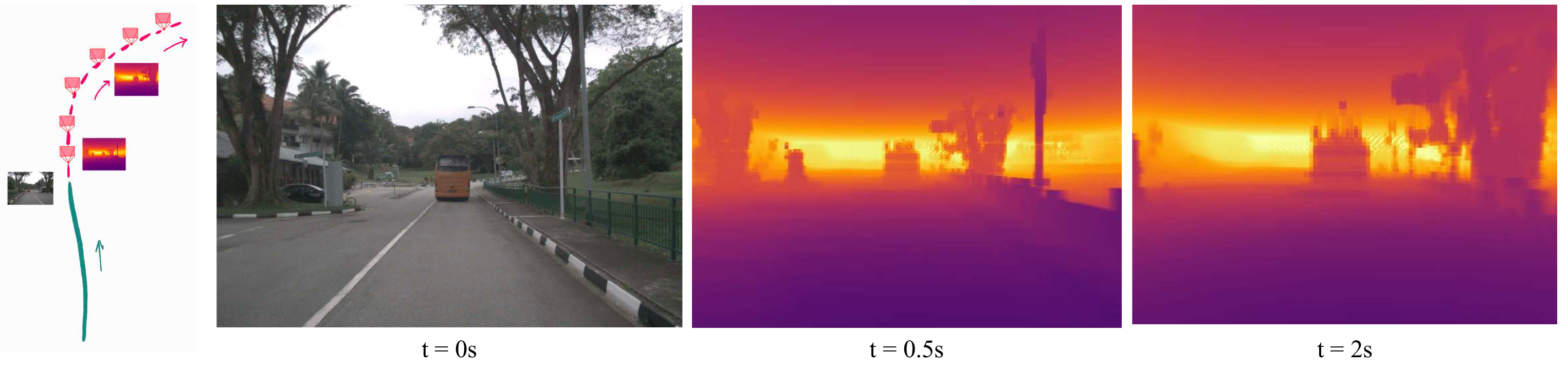}
    \caption{\textbf{Novel extrinsic-view synthesis} Dense depth maps rendered from the predicted future 4D occupancy from novel viewpoints. To render these depth maps, we take a novel future trajectory of the egovehicle. Placing the camera at each of these locations, always facing forward into the voxel grid (shown in the future dotted red trajectory on the left), gives us a camera coordinate system in which we can shoot rays from the camera center to every pixel in the image, and further beyond into the 4D occupancy volume. Every pixel represents the expected depth along its ray. The RGB image at $t=0s$ is shown as reference and is not used in this rendering. For the depth maps, darker is closer, brighter is farther. Depth on sky regions is untrustworthy as no returns are received for this region from the LiDAR sensor.}
    \label{fig:densify}
\end{figure*}

The main observation is that the static variant is a powerful baseline for short-horizon forecasting. This is because a single voxel grid serves as a dense static map of the local region, and since an extremely high majority of the world remains static, this is expected to be a reasonable baseline for short-horizon forecasting. Note that this variant is still stronger than the ray tracing baseline in Tab. \ref{tab:reeval} because of its ability to hallucinate occluded parts of the world. On the other hand, the proposed \textit{dynamic} variant (which predicts one voxel grid per future timestep), performs the best at long-horizon forecasting. With the residual variant, our hope was to separate dynamic scene elements from static regions, but in practice this decomposition fails as there is not enough regularization to force motion-based separation.

Since, the static variant outperforms the state-of-the-art on 1s forecasting, we analyse these variants further in the appendix by using the segmentation annotations on nuScenes-LiDARSeg \cite{lidarseg} and computing the proposed metrics separately on foreground and background points. This helps us understand which regions in the scene contribute the least to the performance of these variants.

\subsection{Applications}

\paragraph{Generalization across sensors}

In Fig.~\ref{fig:domaintransfer} (captioned as new intrinsic-view synthesis), we show how one can render point clouds as if they are captured by different LiDAR sensors from the same predicted future occupancy. Typically, different LiDAR sensors exhibit different ray patterns when sensing. %
For the case shown, the nuScenes LiDAR is an ``in-domain" sensor, i.e., the occupancy grid was predicted by a network learned over LiDAR sweeps captured by a nuScenes LiDAR. The KITTI and ArgoVerse LiDARs are ``out-of-domain". We hope that learning of such a generic representation allows methods in sensor domain transfer \cite{yi2021complete, langer2020domain} to look at the task from the perspective of spacetime 4D occupancy. The formulation we have laid out also makes it easy to train across different datasets, making zero-shot cross-dataset transfer possible for LiDARs \cite{ranftl2020towards, ranftl2021vision}. In the previous section and in Tab. \ref{tab:kittiodo}, we highlight the first result in this direction, where our method trained on the ArgoVerse2.0 dataset when tested on KITTI-Odometry beats the prior art \cite{mersch2022self} on KITTI-Odometry. Furthermore, our proposed disentangling also allows for multi-dataset training, for which we point the readers to the appendix.
\vspace{-0.3cm}
\paragraph{Novel view synthesis}
In Fig.~\ref{fig:densify} (captioned as new-extrinsic view synthesis), we show dense depth maps rendered from our learnt occupancy grid using novel ego-vehicle trajectories or viewpoints. Such dense depth of a scene is not possible to get from existing LiDAR sensors that return sparse observations of the world. Although classical depth completion \cite{Uhrig2017THREEDV} from sparse LiDAR input exists as a single-frame (current timestep) task, here we note that with our representation, it is possible to densify sparse LiDAR point clouds from the \textit{future}, with such rendered depth maps backprojected into 3D. This dense 360$^{\circ}$ depth is evaluated on sparse points (with the help of future LiDAR returns) by our proposed ray-based evaluation metrics.

\section{Conclusion}
In this paper, we propose looking at point cloud forecasting through the lens of geometric occupancy forecasting, which is an emerging self-supervised task \cite{khurana2022differentiable}, originally set in the birds'-eye-view but extended to full 3D through this work. We advocate that this shift in viewpoint is necessary for two reasons. First, this shift helps algorithms focus on a generic intermediate representation of the world, i.e. its spatiotemporal 4D occupancy, which has great potential for downstream tasks. Second, this ``renovates'' how we formulate self-supervised LiDAR point cloud forecasting \cite{mersch2022self, weng2022s2net, weng2021inverting} by factoring out sensor extrinsics and intrinsics from the learning of shape and motion of different scene elements. In the end, we reiterate that the two tasks in discussion are surprisingly connected. We propose an evaluation protocol, that unifies the two worlds and focuses on a scalable evaluation for predicted geometry.

\newpage

\appendix
\section*{\centering Appendix}
In this supplement, we extend our discussion of the proposed reformulation of point cloud forecasting into 4D occupancy forecasting. Specifically, we discuss details about the network architecture of the proposed approach in Sec. \ref{sec:arch}, further results on nuScenes, KITTI-Odometry and ArgoVerse2.0 in Sec. \ref{sec:SOTA}, and the quality of our forecasts separately on the foreground and background points in Sec. \ref{sec:fgbg}.

\section{Network details}
\label{sec:arch}

\paragraph{Architecture implementation}
We build on top of the encoder-decoder architecture first proposed by Zeng \etal~\cite{zeng2019end} for neural motion planning. We extend the version of this architecture used by Khurana \etal~\cite{khurana2022differentiable} for forecasting occupancy in the birds'-eye-view. The only difference between our setup and that used in prior work~\cite{khurana2022differentiable}, is that we treat our 4D voxel grid (X x Y x Z x T) as a reshaped 3D voxel grid (X x Y x ZT), where the Z or height dimension is incorporated into the channel dimension of the input, allowing us to still make use of 2D convolutions on a 4D voxel grid. This means that every channel in the input, represents a slice of the world through the height and time dimensions.

\paragraph{Differentiable renderer} We extend the differentiable raycaster developed by Khurana \etal~\cite{khurana2022differentiable} to 3D and employ it as the differentiable voxel renderer in our approach. As in the prior work, we define our set of rays using the position of the egovehicle in the global coordinate frame as the origin, and all the LiDAR returns as the end points for the rays. The 4D voxel grid is initialized with three labels - empty, occupied and unknown based on the returns in the LiDAR sweeps. Each ray is traversed using a fast voxel traversal algorithm \cite{amanatides1987fast}. Given all the voxels and their occupancies along a ray, we compute the expected distance the ray travels through the voxel grid. This is same as volume rendering but in a discretized grid ~\cite{mildenhall2020nerf}. The gradient of the loss between this expected distance and the groundtruth distance is backpropagated to all the voxels traversed by the ray. Note that when a ray does not terminate within the voxel grid volume, we put all the probability mass of occupancy at the boundary of the voxel grid, similar to Mildenhall \etal~\cite{mildenhall2020nerf}. This means that when a ray passes through occupancy regions that are empty (refer occupancy visuals in the main draft and summary video), the rays results in a point at the boundary of the voxel grid.

\paragraph{Dataset training and testing splits} We use the official train and validation splits of nuScenes and ArgoVerse2.0. Only when comparing results on KITTI-Odometry with ST3DCNN \cite{mersch2022self}, we follow their dataset splits for training and testing. These dataset splits allow us to draw apples-to-apples comparisons with state-of-the-art approaches.

\section{Additional results}
\label{sec:SOTA}

\subsection{nuScenes}

\begin{figure*}
    \centering
    \includegraphics[width=\linewidth]{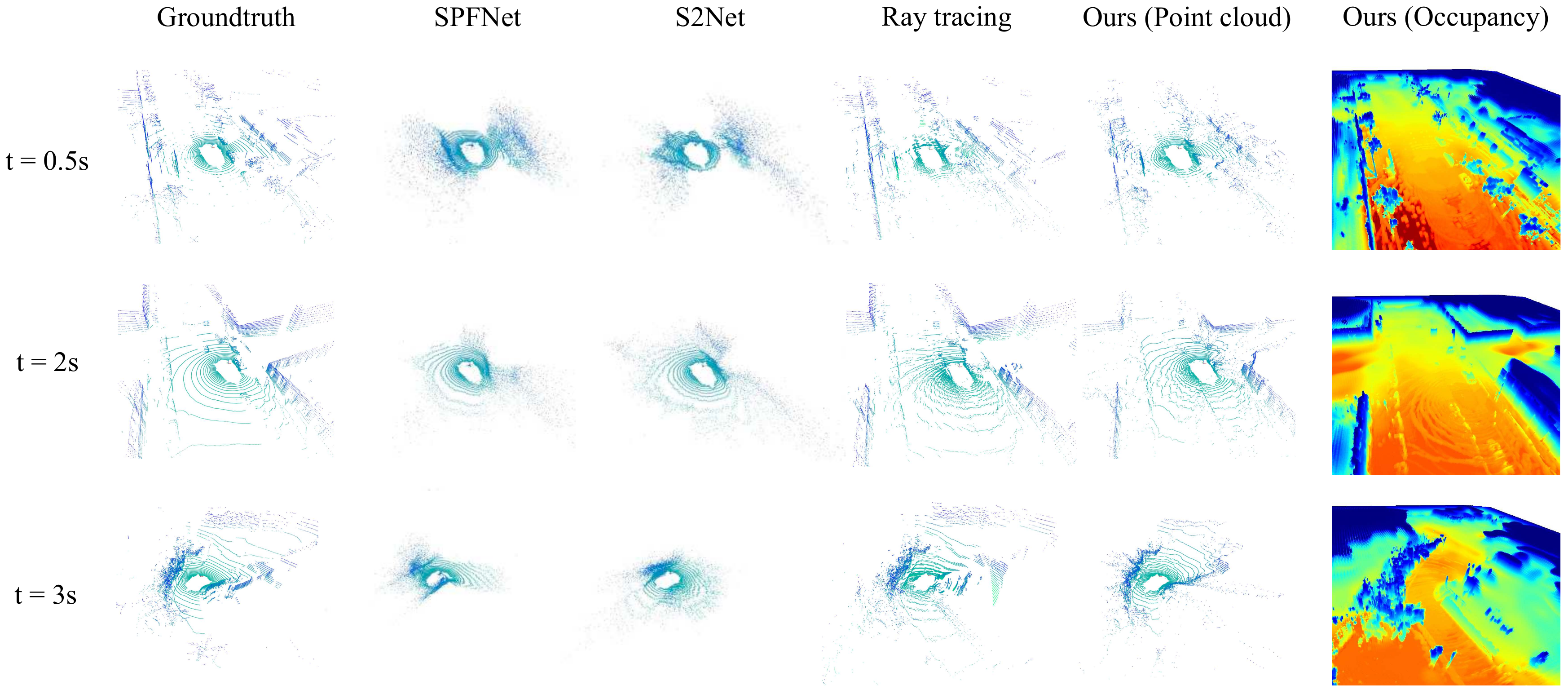}
    \caption{Qualitative results on the nuScenes dataset on three different sequences at different time horizons. We compare the point cloud forecasts of our approach with the aggregation-based ray tracing baseline and S2Net~\cite{weng2022s2net}, SPFNet~\cite{weng2021inverting} with confidence filtering, after applying a recommended confidence threshold of 0.05 on the point clouds. Our forecasts look significantly crisper than the SOTA, however we see that the ray tracing baseline is also a strong baseline.  We visualize a render of the learnt occupancy and the color encodes height along the z-axis.
    \vspace{1cm}}
    \label{fig:maskedeval}
\end{figure*}

\paragraph{Results with confidence thresholding} We supplement the results in the main paper by evaluating the point cloud forecasts of SPFNet \cite{weng2021inverting} and S2Net \cite{weng2022s2net} by thresholding points at a recommended confidence threshold of 0.05. Qualitatively in Fig. \ref{fig:maskedeval}, we observe point clouds from SOTA that only consist of high confidence LiDAR returns close to the ground plane, because of which we perform quantitatively much better than these baselines on our ray-based metrics. We summarise these results in Tab.~\ref{tab:maskedeval}.

\paragraph{Access to ground-truth egoposes during evaluation} Note that in our proposed formulation of 4D occupancy forecasting, we view the LiDAR point clouds used during training as just another observation of the world, which in our case, happens to come from the view of the ego-vehicle. In reality, this LiDAR measurement of occupancy could have also come from any other observer in the world. Similarly, during evaluation, the only LiDAR measurement we have access to comes from the view of the ego-vehicle, making this the only datapoint to evaluate our occupancy forecasts against. This creates an apparent advantage for our method when comparing to point cloud forecasting approaches because they do not have access to ground-truth egoposes from the future. To alleviate this concern, first, we use the future ground-truth egoposes to align all point cloud forecasts to a global coordinate frame. Only after doing this, all the reported metrics are computed for the baselines. Second, we employ a simple motion planner based on linear dynamics, and use these planned future egoposes for evaluating our own method. We see that the metrics drop marginally, showing that the dependence of our method on ground-truth egoposes from the future is not a concern. This is also true for the ray tracing baseline, results of which are summarised in Tab. \ref{tab:constvel}.

\begin{table}[t]
  \centering
  \resizebox{\columnwidth}{!}{%
  \begin{tabular}{@{}cccccc@{}}
  \toprule[1.5pt]
	\multirow{2}{*}{Method} & \multirow{2}{*}{Horizon} & \multirow{2}{*}{L1 (m)} & \multirow{2}{*}{AbsRel (\%)} & \multicolumn{2}{c}{Chamfer Distance ($m^2$)}\\
	& & & & Near-field & Vanilla\\
    \midrule
    \multirow{2}{*}{S2Net \cite{weng2022s2net}} & 1s & 2.88 & 20.57 & 4.61 & 11.77 \\
    & 3s & 4.97 & 24.79 & 13.10 & 30.95 \\
    \midrule
    \multirow{2}{*}{SPFNet \cite{weng2021inverting}} & 1s & 5.30 & 30.12 & 21.24 & 45.12 \\
    & 3s & 5.70 & 28.65 & 20.99 & 44.71 \\
        \midrule
    \multirow{2}{*}{Ray tracing} & 1s & 1.50 & 14.73 & 0.54 & \textbf{0.90} \\
    & 3s & 2.44 & 26.86 & 1.66 & \textbf{3.59} \\
    \midrule
        \multirow{2}{*}{Ours} & 1s & \textbf{1.40} & \textbf{10.37} & \textbf{1.41} & 2.81 \\
    & 3s & \textbf{1.71} & \textbf{13.48} & \textbf{1.40} & 4.31 \\
    \bottomrule[1.5pt]
  \end{tabular}
  \caption{Results on nuScenes \cite{caesar2019nuscenes} with confidence filtering on SPFNet and S2Net. As described in the main paper we threshold the points at a recommended confidence threshold of 0.05. We see that the conclusions made from the proposed metrics are more in line with the qualitative results in Fig.~\ref{fig:maskedeval}. This once again reiterates the need for metrics that intuitively evaluate the underlying \textit{geometry} of the scene instead of uncorrelated samples of the scene (e.g., points in space).
  }
  \label{tab:maskedeval}
}
\end{table}

\begin{table}[t]
  \centering
  \resizebox{\columnwidth}{!}{%
  \begin{tabular}{@{}cccccc@{}}
  \toprule[1.5pt]
	\multirow{2}{*}{Method} & \multirow{2}{*}{GT Egoposes} & \multirow{2}{*}{L1 (m)} & \multirow{2}{*}{AbsRel (\%)} & \multicolumn{2}{c}{Chamfer Distance ($m^2$)} \\
	& & & & Near-field & Vanilla \\
     \midrule
    Ray tracing & Yes & 2.44 & 26.86 & 1.66 & 3.59 \\
    Ray tracing & No & 2.50 & 26.35 & 1.60 & \textbf{3.39} \\
    Ours & Yes & \textbf{1.71} & \textbf{13.48} & \textbf{1.40} & 4.31\\
    Ours & No & 1.84 & 13.95 & 1.50 & 4.50\\
    \bottomrule[1.5pt]
  \end{tabular}
  \caption{We experiment with using a simple linear dynamics based motion planner that can replace the ground-truth future egoposes used in our analysis. Our experiments prove that even in the absence of access to ground-truth future egoposes -- which are not a concern from the viewpoint of our formulation but only a means to evaluate the occupancy predictions -- simple linear dynamics models such as those based on constant velocity, suffice.}
  \label{tab:constvel}
  }
\end{table}

\begin{table}[t]
  \centering
  \resizebox{\columnwidth}{!}{%
  \begin{tabular}{@{}cccccc@{}}
  \toprule[1.5pt]
	\multirow{2}{*}{Method} & \multirow{2}{*}{Horizon} & \multirow{2}{*}{L1 (m)} & \multirow{2}{*}{AbsRel (\%)} & \multicolumn{2}{c}{Chamfer Distance ($m^2$)}\\
	& & & & Near-field & Vanilla \\
    \midrule
    \multirow{2}{*}{Ray tracing} & 1s & 2.39 & 15.43 & \textbf{0.56} & \textbf{1.90}\\
    & 3s & 3.72 & 25.24 & 2.50 & \textbf{11.59} \\
    \midrule
    \multirow{2}{*}{Ours} & 1s & \textbf{2.25} & \textbf{10.25} & 1.53 & 60.94 \\
    & 3s & \textbf{2.86} & \textbf{14.62} & \textbf{2.20} & 69.81 \\
    \bottomrule[1.5pt]
  \end{tabular}
  \caption{Quantitative results on the ArgoVerse2.0 \cite{wilson2021argoverse} dataset. We compare our method trained on the ArgoVerse2.0 dataset to the ray tracing baseline and find similar trends to nuScenes and KITTI-Odometry.}
  \label{tab:argoverse}
  }
\end{table}

\subsection{KITTI-Odometry}

\begin{figure*}
    \centering
    \vspace{1cm}
    \includegraphics[width=\linewidth]{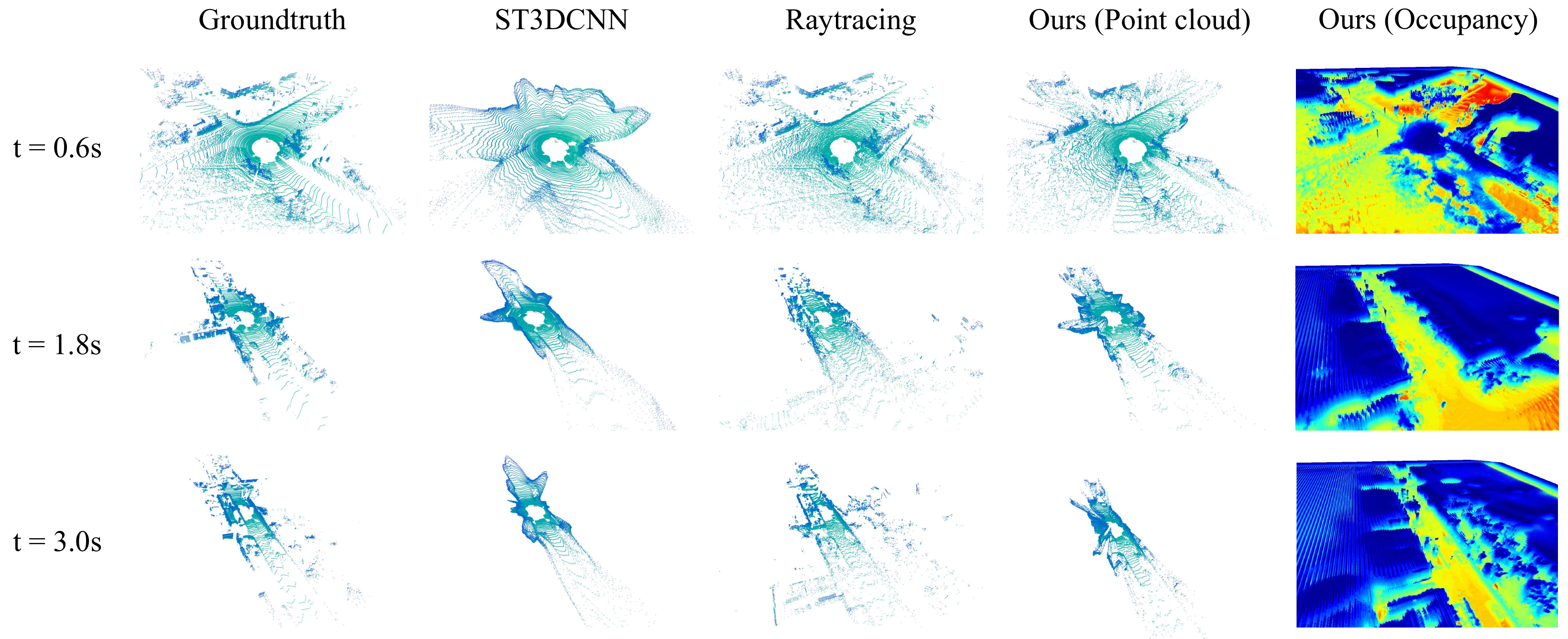}
    \caption{Qualitative results on KITTI-Odometry on three different sequences at different time horizons. We compare the point cloud forecasts of ST3DCNN \cite{mersch2022self} and the ray tracing baseline. We see that this SOTA is qualitatively more geometry-aware than the SOTA on nuScenes. However, our method is still more reflective of the true rigid geometry of the underlying world. We visualize a render of the learnt occupancy and the color encodes height along the z-axis.
    \vspace{0.5cm}
    }
    \label{fig:kittiodo}
\end{figure*}

\paragraph{Qualitative results} We supplement the quantitative results in the main paper with qualitative results in Fig. \ref{fig:kittiodo}. As noted before, the trends are similar to nuScenes.

\subsection{ArgoVerse2.0}

We benchmark ArgoVerse2.0 in Tab.~\ref{tab:argoverse} and compare the ray tracing baseline to our method.
We see that the ray tracing baseline is strong and performs better than our method in terms of the Chamfer distance. Yet, our method is able to forecast better scene geometry in the near-field, than the ray tracing baseline as suggested by the ray-based metrics.

Note that the high vanilla Chamfer distance for ArgoVerse2.0 is due to the fact that the its LiDAR is long-range (up to 200m) and we focus on points only withing the bounded volume. We also clarify that we always test cross-sensor generalization between KITTI-Odometry and ArgoVerse2.0, with training on ArgoVerse2.0 because (1) both datasets have the same number of LiDAR beams and point clouds are captured at the same frequency, and (2) ArgoVerse2.0 is a much larger and diverse dataset than KITTI-Odometry that is suitable for pretraining.

\begin{table*}[t]
  \centering
  \resizebox{\columnwidth}{!}{%
  \begin{tabular}{@{}cccccccccccccccccc@{}}
  \toprule[1.5pt]
    \multirow{3}{*}{Config} & \multirow{3}{*}{Horizon} & \multicolumn{4}{c}{Pedestrians} & \multicolumn{4}{c}{Vehicles} & \multicolumn{4}{c}{All foreground} & \multicolumn{4}{c}{Background}\\
    \cmidrule(l{7pt}r{7pt}){3-6}
    \cmidrule(l{7pt}r{7pt}){7-10}
    \cmidrule(l{7pt}r{7pt}){11-14}
    \cmidrule(l{7pt}r{7pt}){15-18}
	& & \multirow{2}{*}{L1} & \multirow{2}{*}{AbsRel} & \multicolumn{2}{c}{Chamfer Dist.} & \multirow{2}{*}{L1} & \multirow{2}{*}{AbsRel} & \multicolumn{2}{c}{Chamfer Dist.} & \multirow{2}{*}{L1} & \multirow{2}{*}{AbsRel} & \multicolumn{2}{c}{Chamfer Dist.} & \multirow{2}{*}{L1} & \multirow{2}{*}{AbsRel} & \multicolumn{2}{c}{Chamfer Dist.} \\
	& & & & N.f. & Vanilla & & &  N.f.& Vanilla & & &  N.f. & Vanilla & & &  N.f. & Vanilla \\
    \midrule
    Ray tracing & 1s & \textbf{6.09} & 37.18 & \textbf{61.30} & \textbf{66.79} & \textbf{3.53} & 28.82 & \textbf{16.92} & \textbf{21.19} & 3.72 & 34.47 & \textbf{16.09} & 19.45 & 1.39 & 12.51 & \textbf{0.49} & \textbf{0.85} \\
    Ours & 1s & 6.43 & \textbf{34.89} & 79.63 & 68.60 & 3.61 & \textbf{25.28} & 21.47 & 22.59 & \textbf{3.61} & \textbf{28.33} & 18.19 & \textbf{19.05} & \textbf{1.33} & \textbf{8.82} & 1.44 & 3.02 \\
    \midrule
    Raytracing & 3s & 7.84 & 46.42 & 92.86 & 92.97 & 5.29 & 44.25 & 26.99 & 38.22 & 5.52 & 51.48 & 25.66 & 35.32 & 2.27 & 23.50 & 1.60 & \textbf{3.48} \\
    Ours & 3s & \textbf{6.58} & \textbf{34.72} & \textbf{78.47 }& \textbf{71.99} & \textbf{4.11} & \textbf{29.73} & \textbf{22.28} & \textbf{28.36} & \textbf{4.14} & \textbf{33.22} & \textbf{18.59} & \textbf{22.57} & \textbf{1.61} & \textbf{11.48} & \textbf{1.43} & 4.63 \\

    \bottomrule[1.5pt]
  \end{tabular}
  \caption{We extend our metrics analysis to pedestrians, vehicles, movable foreground and static background objects separately. Since, we are not able to compute these category-wise metrics for the state-of-the-art \cite{weng2021inverting, weng2022s2net} due to historical reasons, we do this for the ray tracing baseline. In summary, we find that our method outperforms this otherwise strong baseline at long-horizon forecasting. However, in the short-horizon the ray tracing baseline is a strong one; sometimes doing better and other times performing at par with our proposed method.}
  \label{tab:fgbgraytracing}
  }
\end{table*}

\begin{table}[t]
  \centering
  \resizebox{\columnwidth}{!}{%
  \begin{tabular}{@{}cccccc@{}}
  \toprule[1.5pt]
	\multirow{2}{*}{Arch.} & \multirow{2}{*}{Horizon} & \multirow{2}{*}{L1 (m)} & \multirow{2}{*}{AbsRel (\%)} & \multicolumn{2}{c}{Chamfer Distance ($m^2$)}\\
	& & & & Near-field & Vanilla\\
	\midrule
    \multirow{2}{*}{S} & 1s & 3.28 & 26.14 & 13.52 & 15.35 \\
    & 3s & 4.10 & 32.25 & 19.25 & 23.05 \\
    \midrule
    \multirow{2}{*}{D} & 1s & 3.61 & 28.33 & 18.19 & 19.05 \\
    & 3s & 4.14 & 33.22 & 18.59 & 22.57\\
    \midrule
    \multirow{2}{*}{S + R} & 1s & 3.28 & 25.34 & 13.95 & 15.01 \\
    & 3s & 4.11 & 31.65 & 19.91 & 25.29 \\
    \bottomrule[1.5pt]
  \end{tabular}}
  \caption{Performance analysis on \textbf{foreground} query rays on nuScenes \cite{caesar2019nuscenes} for the different architecture variants introduced in the main draft, static (S), dynamic (D) and residual (S+R).}
  \label{tab:fg}
\end{table}

\begin{table}[t]
  \centering
  \resizebox{\columnwidth}{!}{%
  \begin{tabular}{@{}cccccc@{}}
  \toprule[1.5pt]
	\multirow{2}{*}{Arch.} & \multirow{2}{*}{Horizon} & \multirow{2}{*}{L1 (m)} & \multirow{2}{*}{AbsRel (\%)} & \multicolumn{2}{c}{Chamfer Distance ($m^2$)}\\
	& & & & Near-field & Vanilla\\
	\midrule
    \multirow{2}{*}{S} & 1s & 1.22 & 7.89 & 1.10 & 3.74 \\
    & 3s & 1.64 & 11.65 & 1.43 & 4.00 \\
    \midrule
    \multirow{2}{*}{D} & 1s & 1.33 & 8.82 & 1.44 & 3.02 \\
    & 3s & 1.61 & 11.48 & 1.43 & 4.63 \\
    \midrule
    \multirow{2}{*}{S + R} & 1s & 1.29 & 8.48 & 1.07 & 3.52 \\
    & 3s & 1.74 & 12.01 & 1.56 & 3.78 \\
    \bottomrule[1.5pt]
  \end{tabular}}
  \caption{Performance analysis on \textbf{background} query rays on nuScenes \cite{caesar2019nuscenes} for the different architecture variants introduced in the main draft, static (S), dynamic (D) and residual (S+R).}
  \label{tab:bg}
\end{table}

\section{Foreground vs. background query rays}
\label{sec:fgbg}

In order to further analyse the variants of our architecture, we separate the query rays as belonging to foreground or background regions, using the labels from nuScenes' LiDARSeg \cite{lidarseg}. We evaluate both the regions using both the new and old metrics in Table \ref{tab:fg} and Table \ref{tab:bg}.

\paragraph{Poor performance on foreground objects} Our main observation is that all the variants perform poorly on the foreground objects (which includes moving or stationary foreground objects) as compared to the background. This is because a large number of rays and voxels (more than 90\%) belong to background regions and thus, the foreground objects are downweighted during the training process. Even when the combined evaluation of foreground and background regions is considered (Table 3 in main draft), we see that the poor performance on the foreground fails to materialize in the metrics. This hints are improving the metrics and methods to focus more on the forecasting of foreground objects, especially those in motion.

\paragraph{Strengths of each variant} Another observation stemming from the above fact is that even with this disentangled evaluation on foreground, the \textit{static} variant is the strongest baseline for short-horizon forecasting (1s). On 3s forecasting, the \textit{dynamic} variant shines on the ray-based evaluation of background objects (some unseen background regions may only appear at future timesteps) and the \textit{residual} variant shines on the ray-based evaluation of foreground objects (possibly decouples the foreground from background regions better).

\paragraph{Comparison to the ray tracing baseline} Given the strength of the ray tracing baseline, we investigate its performance on foreground and background objects in comparison to our approach in Tab. \ref{tab:fgbgraytracing}. This time we further divide foreground objects into subcategories of pedestrains and vehicles, while also reporting the metrics on all foreground objects. Note that the according to the vocabulary of nuScenes, apart from different types of pedestrians and vehicles, miscellaneous movable objects like traffic cones and barriers are included in the umbrella category of foreground objects. We have the following findings:

\begin{enumerate}
    \item For long-horizon forecasting, our method consistently does better than the ray tracing baseline, for both all types of foreground objects and background objects.
    \item For short-horizon forecasting, the ray tracing baseline performs at par with our method and sometimes even better (on most types of foreground objects), hence proving to be a strong yet simple and non-learnable approach that does not require any training data.
\end{enumerate}

{\small
\bibliographystyle{ieee_fullname}
\balance
\bibliography{egbib}
}

\end{document}